\documentclass[12pt]{iopart}

\usepackage{multirow}
\usepackage{graphicx}
\usepackage{subfigure}

\usepackage{natbib}
\usepackage{iopams}

\usepackage{hyperref}
\usepackage{bm}
\expandafter\let\csname equation*\endcsname\relax
\expandafter\let\csname endequation*\endcsname\relax
\usepackage{amsmath}

\begin{document}

\title[Transformer-based Lung Nodule Detection]{Unsupervised Contrastive Learning based Transformer for Lung Nodule Detection}

\author{Chuang Niu$^1$ and Ge Wang$^1$}


\address{$^1$Biomedical Imaging Center, Department of Biomedical Engineering, Rensselaer Polytechnic Institute, Troy, New York, United States of America}
\ead{wangg6@rpi.edu}
\vspace{10pt}

\begin{abstract}
Early detection of lung nodules with computed tomography (CT) is critical for the longer survival of lung cancer patients and better quality of life. Computer-aided detection/diagnosis (CAD) is proven valuable as a second or concurrent reader in this context. However, accurate detection of lung nodules remains a challenge for such CAD systems and even radiologists due to not only the variability in size, location, and appearance of lung nodules but also the complexity of lung structures. This leads to a high false-positive rate with CAD, compromising its clinical efficacy. Motivated by recent computer vision techniques, here we present a self-supervised region-based 3D transformer model to identify lung nodules among a set of candidate regions. Specifically, a 3D vision transformer (ViT) is developed that divides a CT image volume into a sequence of non-overlap cubes, extracts embedding features from each cube with an embedding layer, and analyzes all embedding features with a self-attention mechanism for the prediction. To effectively train the transformer model on a relatively small dataset, the region-based contrastive learning method is used to boost the performance by pre-training the 3D transformer with public CT images. Our experiments show that the proposed method can significantly improve the performance of lung nodule screening in comparison with the commonly used 3D convolutional neural networks.

\end{abstract}

%
%
%
%
%

\section{Introduction}
Global cancer statistics in 2018 indicates that Lung cancer is the most popular, i.e., 11.6\% of the total cases, and the leading cause of cancer death, up to 18.4\% of the total cancer deaths \citep{lungcancer}. Various studies have shown that early detection and timely treatment of lung nodules can improve the 5-year survival rate \citep{blandin2017progress}. Therefore, major efforts have been made on early and accurate detection of lung nodules in different aspects, such as imaging technologies \citep{nlst, niu2022x}, diagnosis workflows \citep{macmahon2005guidelines}, and computer aided detection and computer aided diagnostic systems \citep{messay2010new}. 
Particularly, recent results indicate that computer aided detection/diagnosis (CAD) systems empowered by artificial intelligence (AI) algorithms as the second or concurrent reader can improve the performance of lung nodule detection on chest radiographs \citep{yoo2021ai} and CT images \citep{roos2010computer, prakashini2016role}.

Lung cancer CAD systems usually involve lung region segmentation, nodule candidate generation, nodule detection, benign and malignant nodule recognition, and different types of lung cancer classification.
In recent years, deep learning methods were developed for CAD systems, continuously improving the performance of some or all the key components in a CAD system.
For example, \citep{harrison2017progressive, hofmanninger2020automatic} showed that deep learning methods for CT lung segmentation significantly improved the performance through training the models with a variety of datasets.
Motivated by the progress in deep learning based objection detection \citep{ren2015faster, lin2017focal} in various domains \citep{7961803, 8410588}, the performance of lung nodule candidate generation and detection was significantly improved by adapting advanced object detection algorithms \citep{jaeger2020retina, baumgartner2021nndetection}.
Nevertheless, high false positive rate is still a main challenge for accurate lung nodule detection \cite{pinsky2018false}.
Clearly, a key step for accurate nodule detection is to effectively reduce the false positive rate for nodule candidates.
Recent studies addressed this issue using various techniques, such as, 3D convolutional neural network \citep{7576695}, multi-scale prediction \citep{cheng2019deep, GU2018220}, relation learning \cite{yang2020relational}, multi-checkpoint ensemble \citep{jung2018classification}, multi-scale attention \citep{zhang2022pulmonary}, etc.
After identifying lung nodules, various methods were proposed to further analyze them, i.e., predicting the malignancy \citep{SHEN2017663, ALSHABI2022108309} and sub-types \citep{LIU2018262, yuan2018hybrid} of lung nodules.
It is exciting that adapting emerging techniques in machine learning and computer vision based on the domain knowledge leads to great progress in CAD systems with great potential for clinical translation.

Recently, transformers \cite{NIPS2017_3f5ee243}, originally developed for natural language processing (NLP), have achieved great successes in various tasks of computer vision. The key component of the transformer is the attention mechanism that utilizes global dependencies between input and output.
For the first time, Vision Transformer (ViT) divides an image into a sequence of non-overlap patches, analyzes them as a sequence of elements similar to words, and produces state-of-the-art results demonstrating the effectiveness and superiority in image classification \citep{dosovitskiy2020image}. Since then, ViT has been successfully applied to various other vision tasks including medical imaging \cite{pan3991087multi} and medical image analysis \citep{lyu2021transformer}.
However, the performance of the original ViT relies on a large labelled image dataset including 300 millions images, and the conventional wisdom is that the transformers do not generalize well if they are trained on insufficient amounts of data. Therefore, directly adopting the transformers for CAD systems is not trivial when labeled data are scarce.

Lack of labeled data is a common problem in the medical imaging and many other fields. A most promising direction of deep learning is the so-called unsupervised or self-supervised learning \cite{niu2020gatcluster, niu2021spice, niu2020suppression} that recently achieved remarkable results which even approach the performance of supervised counterparts.
Particularly, unsupervised learning works by pre-training a neural network on a large scale unlabeled dataset to benefit downstream supervised tasks that only offer a limited number of training samples \cite{he2019moco}.
For unsupervised or self-supervised learning, the pretext task is the core to learn meaningful representation features.
Recent progresses suggest that instance contrastive learning \citep{chen2020simple} and masked autoencoding \citep{he2021masked} are two  most effective and scalable pretext tasks for unsupervised representation learning.
Specifically, instance contrastive learning maximizes the mutual information between two random transformations of the same instance (e.g., an object in a natural image or a patient represented by a CT volume).
This can be achieved by forcing the representation features from different transformations of the same instance to be similar while the features from different instances to be dissimilar.
On the other hand, masked autoencoding recovers masked parts from the rest visible data, which has been used for pre-training in various tasks and recently produced encouraging results \citep{he2021masked}, using an asymmetric encoder-decoder architecture and a high proportion masking strategy.

Based on the above progresses, here we study how to effectively adapt ViT and unsupervised pretraining for lung nodule detection, so that the false positive rate can be reduced for lung nodules to be effectively singled out of a set of candidates, in comparison with the commonly used 3D CNNs.
In our work, we adapt the original transformer to a CT image volume with the fewest possible modifications.
Advantages of keeping the original transformer configuration as much as possible include the scalability in the modeling capacity and the applicability across multiple modality datasets.
With this preference in mind, we simply divide an 3D CT image volume into non-overlap cubes and extract their linear embeddings as the input to the transformer.
These cubes are equivalent to the tokens or words in NLP.
However, without pretraining on large-scale datasets, the superiority of the transformer cannot be realized, especially for lung nodule analysis where labeled data are usually expensive and scarce, e.g., there are only over one thousand labels in public datasets.
To overcome this difficulty, we perform unsupervised region-based contrastive learning on public CT images from the LIDC-IDRI dataset to effectively train the adapted transformer. Our experimental results show that while the adapted 3D transformer trained with a relatively small number of labeled lung nodule data from scratch achieved worse results than the 3D CNN model, the pretraining techniques enabled the adapted transformer to outperform the commonly used 3D CNN. Interestingly, we found that unsupervised pretraining is more effective than supervised pretraining with natural images in a transfer learning manner to boost the performance of the adapted transformer. 

The rest of this paper is organized as follows. In the next section, we describe our transformer architecture and implementation details. In the third section, we report our experimental design and representative results in comparison to competing CNN networks. In the last section, we discuss relevant issues and conclude the paper.

\section{Methodology}

\subsection{Vision Transformer for Lung Nodule Detection}
\begin{figure*}[bt!]
    \centering
    \includegraphics[width=1\textwidth]{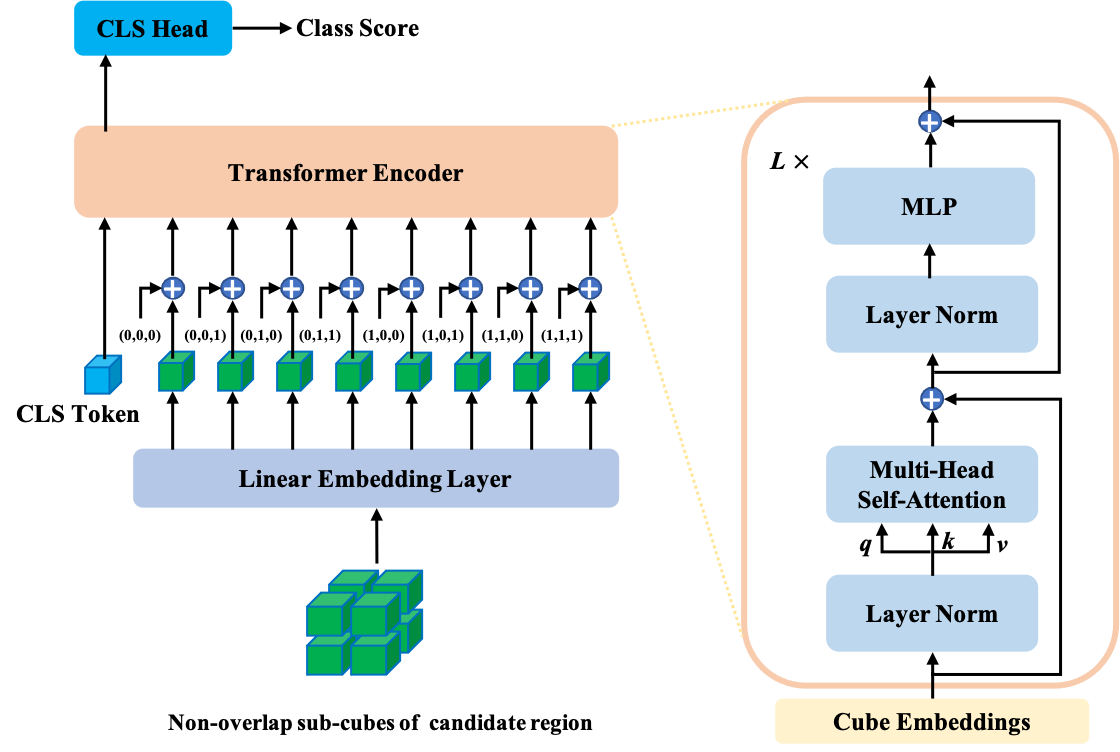}
    \caption{Transformer architecture for lung nodule detection.}
    \label{fig_trans}
\end{figure*}

In this section, we describe the architecture of our adapted Transformer for lung nodule detection in a CT image volume. The whole architecture is depicted in Fig. \ref{fig_trans}, where there are four parts. The details on each part are given as follows.

\noindent \textbf{Input and Linear Embedding:} The input is a 3D tensor, $\bm{x} \in \mathbb{R}^{H\times W \times D}$, which is a local candidate volumetric region of interest in a whole CT volume. Similar to what ViT does, the image volume is divided into a sequence of non-overlap cubes, $\bm{x}_c \in \mathbb{R}^{S\times S \times S}$, similar to words in NLP, where $H, W, D$ are the input volume size, $S$ is the cube size, and $c$ is the index for cubes. Then, the linear embedding layer maps these cubes to embedding features independently. In practice, the linear embedding layer is implemented as a 3D convolutional layer, where both the kernel size and convolutional stride are $S \times S \times S$. Therefore, this embedding layer can directly take the original 3D volume as input and outputs the embedding features of non-overlapped cubes, i.e., $\bm{z_c} = \bm{E}(\bm{x}_c) \in \mathbb{R}^{d}$, where $c = 1, 2, ...,$ 
$S\times S\times S$, 
and $d$ is the dimension of embedding. As in ViT, the $[class]$ token of a learnable embedding is prepended to the sequence of embedded cubes, and the final sequence of linear embedding features are denoted as $[\bm{z}_0; \bm{z}_1; \cdots; \bm{z}_N]$, where $\bm{z}_0$ denotes the learnable class embedding, and $N = S\times S\times S + 1$ is the total number of input embeddings.

\noindent \textbf{Position Embedding:}
For the model to be aware of the relative position of each cube, position embeddings are coupled with the feature embeddings.
In this study, we extend the sin-cosine position encoding \citep{dosovitskiy2020image} into the 3D space.
Specifically, sine and cosine functions of different frequencies are used to encode 3D position information as
\begin{equation}
    \begin{aligned}
        &PE(x,y,z) = [PE_{sin}(x), PE_{cos}(x), PE_{sin}(y), PE_{cos}(y), PE_{sin}(z), PE_{cos}(z)], \\
        &PE_{sin}(p) = sin(p/10000^{i/d_{pos}}), i = 0, 1, \cdots, d_{pos}-1, \\
        &PE_{cos}(p) = cos(p/10000^{i/d_{pos}}), i = 0, 1, \cdots, d_{pos}-1, 
    \end{aligned}
\end{equation}
where $(x, y, z)$ is the relative position of a cube and $PE(x, y, z) \in \mathbb{R}^{d}$ is the corresponding position embedding, here the position embedding of the class token is a zero vector. The position embedding consists of six parts and the dimension of each part is $d/6$.
To be consistent with the notations of feature embeddings, we use $PE_c$ denotes the position embedding of a specific cube.
Finally, the point-wise summations of position and feature embeddings are input to the transformer encoder.

\noindent \textbf{Transformer Encoder:}
The transformer encoder consists of L stacked identical blocks, where each block has two layers, i.e., a multi-head self-attention layer and a simple positionwise fully-connected layer.
As shown in Fig.~\ref{fig_trans}, the residual connection and layer normalization are applied in these two sub-layers.
More specifically, given a sequence of input embeddings, $\bm{z}^0 = [\bm{z}_0^0 + PE_0; \bm{z}_1^0 + PE_1; \cdots, \bm{z}_{S^3}^0 + PE_{S^3}] \in \mathbb{R}^{N\times d}$, the output of the $l^{th}$ multi-head self-attention layer is computed as
\begin{equation}
    \begin{aligned}
        &[\bm{q}^{lm}, \bm{k}^{lm}, \bm{v}^{lm}] = LN(\bm{z}^{l-1})\bm{U}^{lm}_{qkv}, \\
        &\bm{A}^{lm} = softmax(\bm{q}^{lm}{\bm{k}^{lm}}^{T}), \\
        &\bm{z}^{lm} = \bm{A}^{lm}\bm{v}^{lm}, m = 1, 2, \cdots, M, \\
        & \bm{z}^l_{att} = [\bm{z}^{l1}, \bm{z}^{l2}, \cdots, \bm{z}^{lM}]\bm{U}^l_{msa} + \bm{z}^{l-1}, l = 1, 2, \cdots, L,
    \end{aligned}
\end{equation}
where the first three equations describe the operation of a specific self-attention head and the last equation represents the integration of multiple heads. Specifically, $LN(\cdot)$ denotes the layer norm function, $\bm{U}^{lm}_{qkv} \in \mathbb{R}^{d\times 3d_m}$ represents a linear layer that maps each input embedding vector $\bm{z}^{l-1}$ into three vectors, $\bm{q}^{lm}, \bm{k}^{lm}, \bm{v}^{lm} \in \mathbb{R}^{N \times d_m}$, which are known as the query, key, and value vectors respectively, $\bm{A}^{lm} \in \mathbb{R}^{N\times N}$ is the self-attention weight matrix computed as the inner product between query and key vectors followed by a softmax function. 
Then, the output, $\bm{z}^{lm} \in \mathbb{R}^{N \times d_m}$, of each self-attention head is the weighed sum over all input embeddings to realize a global attention. 
There are $M$ self-attention heads running in parallel, with $m$ being the head index, which jointly attend to information from different representation subspaces at different positions \citep{NIPS2017_3f5ee243}.
To avoid increasing the number of parameters, the vector dimension in each self-attention head is split to $d_m = d / M$.
The output $\bm{z}^l_{att}$ of the multi-head self-attention layer is the concatenation of all self-attention outputs transformed by a linear layer $\bm{U}^l_{msa} \in \mathbb{R}^{d \times D}$ and increased by the signal from the residual connection. Then, this output is forwarded to the MLP layer for the final output of the $l^{th}$ block:
\begin{equation}
    \bm{z}^l = MLP(LN(\bm{z}^l_{att})) + \bm{z}^l_{att}.
\end{equation}
Thus, the final output of the transformer encoder is $\bm{z}^L \in \mathbb{R}^{N \times d}$, which has the same dimension as the input embeddings. 

\noindent \textbf{Classification Head:}
The classification head is a linear layer that projects extracted features by the transformer encoder to classification scores. The classification head only takes the feature vector at the position of the $[class]$ token and outputs a classification score as
\begin{equation}
    \bm{y} = \bm{z}^L_0 \bm{U}_{cls},
\end{equation}
where $\bm{z}^L_0 \in \mathbb{R}^{1 \times d}$, $\bm{U}_{cls} \in \mathbb{R}^{d \times C}$, and $C$ is the number of classes.

\subsection{Region-based Contrastive Learning}

\begin{figure*}[bt!]
    \centering
    \includegraphics[width=1\textwidth]{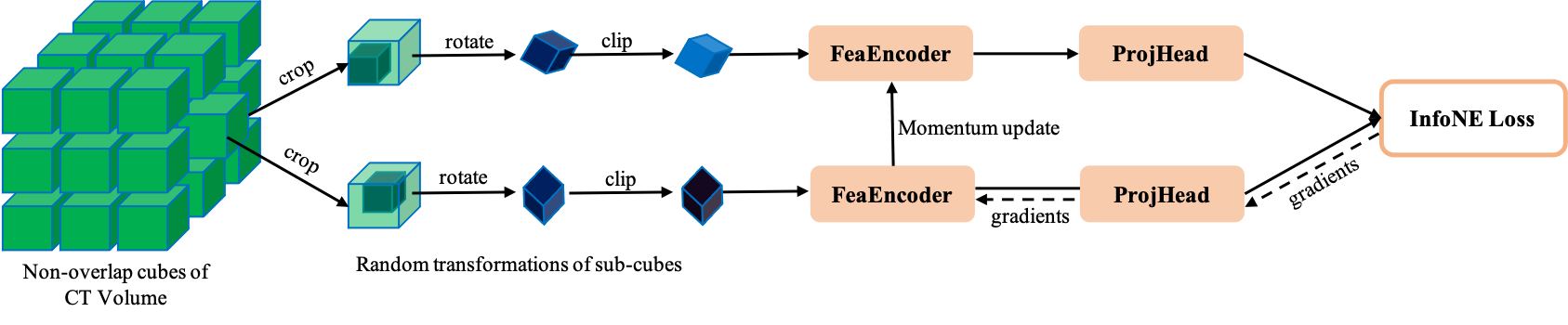}
    \caption{Region-based contrastive learning framework for lung nodule detection.}
    \label{fig_contrast}
\end{figure*}

It is well known that the transformer is extremely data-hungry but labeled lung nodule data is relatively scarce. Hence, we propose a region-based contrastive learning method to pretrain the adapted transformer model by leveraging more unlabeled CT image volumes.
The popular constrastive learning framework is adopted in Fig.~\ref{fig_contrast}.
It consists of two branches that take many pairs of similar samples and outputs the their features.
Generally, this framework enforces similar samples to be closer to each other while dissimilar samples to be more distinct in the representation feature space as measured by the InfoNCE loss.

In the unsupervised context, how to properly define similar and dissimilar samples is the key component \citep{tian2020makes}.
Although great progresses were reported by introducing various random transformation techniques, it is still an open problem on how to keep useful information and compress noise and artifacts in representation features for downstream tasks.
Actually, knowledge on specific downstream tasks plays an important role in defining appropriate similar samples \citep{tian2020makes}.

In our application, similar and dissimilar samples can be defined as follows.
First, as we focus on classifying sub-volumes of a CT volume as lung nodule or not, we divide the whole CT volume into a set of non-overlap cubes and regard each as an unique instance.
This assumes that every 3D sub-region in a patient is different from the others.
Second, two sub-regions with a large intersection should be similar to each other.
Third, although different organs/tissues are usually inspected under different HU windows, the same region under slightly different HU windows should be similar to each other.
Fourth, two sub-regions different by a random rotation should be similar to each other as the angular information is not critical in detecting lung nodules.

Based on the above assumptions, we first divide a CT volume into a set of non-overlap $S_1 \times S_1 \times S_1$ cubes from all patient CT scans to build the whole training dataset, $\{\bm{x}_i\}_{i=1}^{I}$, where $I$ is the total number of cubes.
During training, two sub-cubes of $S_2 \times S_2 \times S_2$ ($S_2 < S_1$) voxels are randomly cropped from a given cube and randomly rotated, and their HU values are randomly clipped, as shown in Fig. \ref{fig_contrast}.
In each training iteration, a set of $B$ cubes are randomly selected, and then each cube is randomly transformed to two sub-cubes $\bm{x}_i', \bm{x}_i''$. Finally, the network parameters are optimized with the InfoNCE loss as follows:
\begin{equation}
    \begin{aligned}
        & \mathcal{L} = \frac{1}{2B}\sum_{i=1}^B(\mathcal{L}(\bm{x}'_i, \bm{x}''_i) + \mathcal{L}(\bm{x}''_i, \bm{x}'_i)), \\
        & \mathcal{L}(\bm{x}'_i, \bm{x}''_i) = - \log\left(\frac{\exp(\mathcal{P}(\mathcal{F}(\bm{x}'_i; \bm{\theta}_{\mathcal{F}}); \bm{\theta}_{\mathcal{P}})^T \mathcal{P}(\mathcal{F}(\bm{x}''_i; \bm{\theta}_{\mathcal{F}}^m); \bm{\theta}_{\mathcal{P}}^m) /\tau)}{\sum_{j=1, j\ne i}^{I} \exp(\mathcal{P}(\mathcal{F}(\bm{x}'_i; \bm{\theta}_{\mathcal{F}}); \bm{\theta}_{\mathcal{P}})^T \mathcal{P}(\mathcal{F}(\bm{x}''_j; \bm{\theta}_{\mathcal{F}}^m); \bm{\theta}_{\mathcal{P}}^m) /\tau)}\right),
    \end{aligned}
\end{equation}
where $\mathcal{F}$ and $\mathcal{P}$ represent the feature encoder and projection head functions with parameters $\bm{\theta}_{\mathcal{F}}$ and $\bm{\theta}_{\mathcal{P}}$ to be optimized, and $\bm{\theta}'_{\mathcal{F}}$ and $\bm{\theta}'_{\mathcal{P}}$ are the moving averaging versions of $\bm{\theta}_{\mathcal{F}}$ and $\bm{\theta}_{\mathcal{P}}$. 
Note that the feature encoder is exactly the transformer model without the classification head, and the projection head is the same as in Moco\_v3.
The loss term $\mathcal{L}(\bm{x}'_i, \bm{x}''_i)$ maximizes the feature similarity between two random transformations from the same cube while minimizing the feature similarity from different cubes. The final loss $\mathcal{L}$ is the average over a batch of $B$ samples.

\subsection{Implementation Details}
\label{sec_implement}
In our adapted transformer, the size of a candidate region was set to $H=W=D=72$, the size of each non-overlap cube $S=8$, the embedding dimension $d=384$, the number of blocks $L=11$, and the number of attention heads $M=12$.
In our region-based contrastive learning, the size of non-overlap sub-regions was set to $S_1=96$, and the size of each input cube $S_2=72$, the low and high HU values of the clip window were randomly sampled from $[-1200, -1000]$ and $[600, 800]$ respectively.
During unsupervised pre-training, the batch size was set to $B=1024$, Adamw was used to optimize the model, the learning rate was $0.0001$ with cosine annealing.
At the fine-tuning stage, only the pretrained linear embedding layer and transformer endoer were kept, the projection head were removed, and a randomly initialized classification head was added, the batch size was set to $64$, and all other hyper parameters for training were kept the same as those in Moco\_v3.
To address the imbalance issue, we randomly sampled each training batch according to the pre-defined positive sampling ratio meaning so that each batch approximately had a fixed ratio of positive to negative samples.
By default, the positive sampling ratio was set to 0.2.

\section{Experimental Design and Results}

\subsection{Dataset and Preprocessing}
\begin{figure*}[bt!]
    \centering
    \includegraphics[width=1\textwidth]{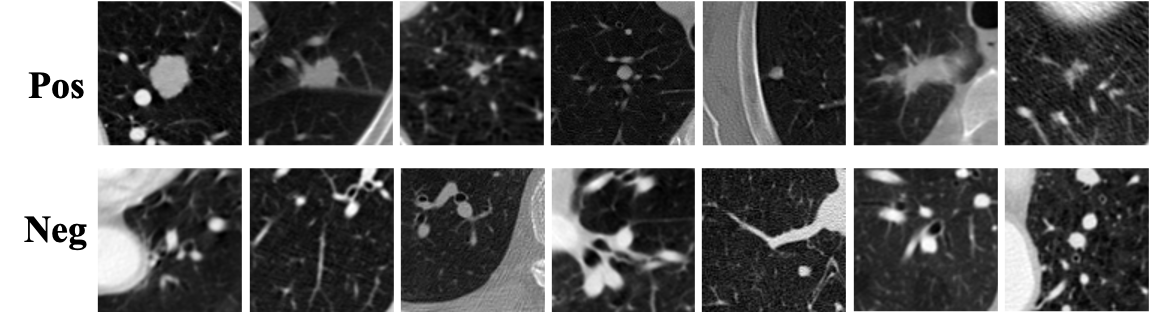}
    \caption{Samples of our lung nodule dataset. The first and second rows show positive and negative samples respectively.}
    \label{fig_sample}
\end{figure*}
In this study, 684 lung nodules of larger than 6mm in size were selected from 436 patients in the LIDC-IDRI \citep{armato2011lung} dataset. As negative samples, $100 \times 436$ (100 from each patient) regions not overlapping with positive regions were randomly selected from the LUNA16 \cite{setio2017validation} dataset. The constructed dataset was divided to a training dataset of 349 patients and a test dataset of 87 patients respectively. All the CT volumes were interpolated along the longitudinal direction so that the longitudinal resolution is the same as the axial resolution.
For region-based contrastive learning, 84,875 non-overlap regions were collected from 801 CT volumes in the LINA16 dataset without any region in the testing dataset.
The region size was set to $96 \times 96 \times 96$ for both supervised and unsupervised learning.
Some positive and negative samples are shown in Fig. \ref{fig_sample}, where it can be seen that the appearance of positive nodules may be very different, and similar structures in the negative regions present strong interferences.
To test the generalizability of different methods, the test set of LUNGx \citep{kirby2016lungx} was used to evaluate the performance of different models trained on the LIDC-IDRI dataset. The LUNGx test set consists of 73 CT scans, each of them contains 1 or 2 positive nodules and 200 negative nodules per scan, and the prepossessing procedure is the same as that for LIDC-IDRI dataset.
Also, these datasets are extremely unbalanced (\#positive:\#negative $\approx$ 1:100 and 1:200), making this task challenging.

\subsection{Evaluation Metric}
Due to imbalance of positive and negative samples in the evaluation dataset, the common Free Response Receiver Operating Characteristic (FROC) curve and Competition Performance Metric (CPM) were used to evaluate the model performance.
Specifically, the true positive rate (TPR) and false positive rate (FPR) are defined as
\begin{equation}
    \begin{aligned}
        TPR &= \frac{TP}{TP + FN}, \\
        FPR &= \frac{FP}{TN + FP},
    \end{aligned}
\end{equation}
where TP, FN, TN, FP are the number of true positive, false negative, true negative, and false positive, respectively.
Then, the average number of false positives per scan, FPS, is defined as
\begin{equation}
    FPS = \frac{FPR \times TN}{NS},
\end{equation}
where NS is the number of CT scans. The FROC curve is plotted as TPR v.s. FPS, which is a variant of the ROC curve, i.e., TPR v.s. FPR.
The CPM score is defined as the average TPR (also called sensitivity) at the predefined FPS points: 0.125, 0.25, 0.5, 1, 2, 4, and 8 respectively.

\subsection{Comparative Analysis}

\begin{figure*}[bt!]
    \centering
    \includegraphics[width=0.7\textwidth]{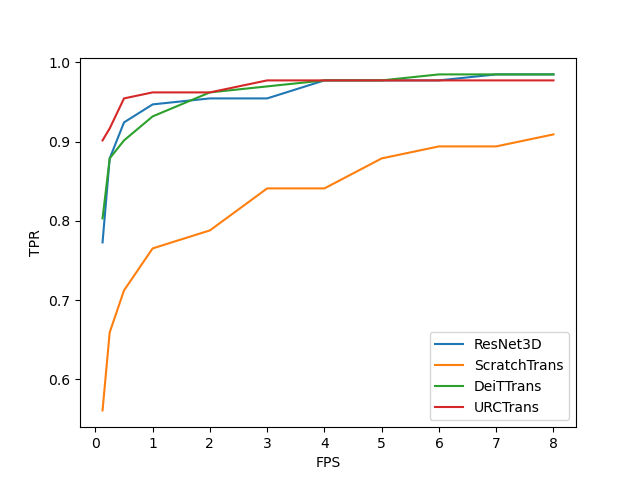}
    \caption{FROC curves of the selected competing methods.}
    \label{fig_froc}
\end{figure*}

\begin{table}[htp]
  \renewcommand{\arraystretch}{1.5}
  \renewcommand\tabcolsep{8pt}
 \caption{Quantitative results. The sensitivities at different FPS points and CPM scores were computed, with the best result highlighted in \textbf{bold}.}
  \centering
  \begin{tabular}{c|ccccccc|c}
    Methods          & 0.125 & 0.25 & 0.5 & 1 & 2 & 4 & 8 & CPM \\
    \hline
    ResNet3D         & 0.773 & 0.879 & 0.924 & 0.947 & 0.955 & 0.977 & 0.985 & 0.920 \\
    ScratchTrans     & 0.561 & 0.659 & 0.712 & 0.765 & 0.788 & 0.841 & 0.909 & 0.748 \\
    DeiTTrans        & 0.803 & 0.879 & 0.902 & 0.932 & 0.962 & 0.977 & 0.985 & 0.920 \\
    URCTrans         & 0.902 & 0.917 & 0.955 & 0.962 & 0.962 & 0.977 & 0.977 & \textbf{0.950} \\
    \hline
  \end{tabular}
  \label{tab:compare}
\end{table}

In this sub-section, we evaluated the effectiveness of the proposed method relative to the following three baselines.
First, we modified ResNet \cite{He_2016_CVPR} to the 3D version, named ResNet3D, as a strong baseline method.
Second, the transformer model was trained from scratch, named ScratchTrans. To make sure that ScratchTrans be sufficiently trained, we doubled the number of training epochs and observed the essentially same results.
Third, we initialized the transformer model with the weights of the pretrained DeiT on the labeled ImageNet dataset in a transfer learning manner, named the resultant network DeiTTrans.
Finally, our proposed transformer model pretrained via unsupervised region-based contrastive learning is referred to as URCTrans.

The comparison results in terms of the sensitivity and CPM scores are summarized in Table \ref{tab:compare}, and the FROC curves are plotted in \ref{fig_froc}. These results show that the ResNet3D model is a very strong baseline with a 0.920 CPM score. The ScratchTrans achieved the worst results among these methods, which is consistent to the results in other domains showing that the transformer is extremely data-hungry and cannot perform well without a large-scale dataset.
Through transfer learning, DeiTTrans significantly improved the performance of the transformer model and produced results similar to that obtained with ResNet3D.
In contrast, our pretraining method without leveraging any labeled data offered the best performance among all comparison methods (0.950 CPM score, 3\% higher than DeiTTrans and the commonly used ResNet3D).
Further inspecting the sensitivities at different FPS points, it can be seen that the URCTrans model performed significantly better than the others when the average number of false positive nodules per scan is small ($\le 1$).
That is the most desired result to effectively avoid falsely reported nodules.
Clearly, our experimental results demonstrate that the transformer pretrained with more CT data through contrastive learning promises a performance superior to the commonly used 3D CNN models.

\begin{table}[htp]
  \renewcommand{\arraystretch}{1.5}
  \renewcommand\tabcolsep{8pt}
 \caption{Generalizability performance results on LUNGx. The sensitivities at different FPS points and CPM scores were computed, with the best result highlighted in \textbf{bold}.}
  \centering
  \begin{tabular}{c|ccccccc|c}
    Methods          & 0.125 & 0.25 & 0.5 & 1 & 2 & 4 & 8 & CPM \\
    \hline
    ResNet3D         & 0.712 & 0.767 & 0.808 & 0.890 & 0.918 & 0.959 & 0.973 & 0.861 \\
    ScratchTrans     & 0.630 & 0.685 & 0.740 & 0.863 & 0.863 & 0.904 & 0.932 & 0.802 \\
    DeiTTrans        & 0.781 & 0.822 & 0.877 & 0.890 & 0.904 & 0.904 & 0.904 & 0.869 \\
    URCTrans         & 0.822 & 0.849 & 0.918 & 0.945 & 0.959 & 0.959 & 0.959 & \textbf{0.916} \\
    \hline
  \end{tabular}
  \label{tab:general}
\end{table}

The generalizability performance results on LUNGx are reported in Table \ref{tab:general}, where the models trained on LIDC-IDRI were directly evaluated.
Although the relative performance of different methods is the same as above, the performance improvement of URCTrans is significantly increased imcomparison with ResNet3D and DeiTTrans counterparts especially for the lower false positive number ($\le 1$).
These results further demonstrated the superiority of the presented method in terms of the generalizability.

\subsection{Effects of the Input Size}

\begin{table}[htp]
  \renewcommand{\arraystretch}{1.5}
  \renewcommand\tabcolsep{8pt}
 \caption{Quantitative results obtained by URCTrans with different input sizes. The best result is highlighted in \textbf{bold}.}
  \centering
  \begin{tabular}{c|ccccccc|c}
    Input size          & 0.125 & 0.25 & 0.5 & 1 & 2 & 4 & 8 & CPM \\
    \hline
    64         & 0.826 & 0.879 & 0.917 & 0.947 & 0.962 & 0.977 & 0.985 & 0.926 \\
    72         & 0.902 & 0.917 & 0.955 & 0.962 & 0.962 & 0.977 & 0.977 & \textbf{0.950} \\
    80         & 0.916 & 0.916 & 0.932 & 0.939 & 0.962 & 0.977 & 0.977 & 0.946 \\
    \hline
  \end{tabular}
  \label{tab:size}
\end{table}

As mentioned in \citep{cheng2019deep, GU2018220}, different sizes of an input tensor allow various levels of contextual information, leading to different performance metrics. Combining the results from multi-scale inputs would boost the performance further.
In this sub-section, we investigate the effect of the input size on the performance of the transformer model for lung nodule detection. The results are in Table~\ref{tab:size}, showing that the medium input size of 72 achieved the best result. It seems heuristic that there is a trade-off between the input size and the model performance, since a too large input may bring more interfering structures while a too small input may not contain enough contextual information to identify lung nodules. Nevertheless, these relatively similar results indicate that the transformer model is robust to the input size.

\subsection{Effects of the Positive Sampling Ratio}

\begin{table}[htp]
  \renewcommand{\arraystretch}{1.5}
  \renewcommand\tabcolsep{5pt}
 \caption{Quantitative results obtained by URCTrans using different positive sampling ratios. The best result is highlighted in \textbf{bold}. The numbers of positive nodules predicted by the transformer model trained with different positive sampling ratios, where the input region is regarded as positive if the prediction score $\ge 0.5$.}
  \centering
  \begin{tabular}{c|ccccccc|c|c}
    Positive ratio          & 0.125 & 0.25 & 0.5 & 1 & 2 & 4 & 8 & CPM & \#Predicted \\
    \hline
    0.1        & 0.795 & 0.841 & 0.894 & 0.924 & 0.933 & 0.947 & 0.969 & 0.900 & 158 \\
    0.2        & 0.826 & 0.879 & 0.917 & 0.947 & 0.962 & 0.977 & 0.985 & \textbf{0.926} &217 \\
    0.3        & 0.765 & 0.856 & 0.917 & 0.932 & 0.969 & 0.977 & 0.985 & 0.915 & 247 \\
    0.4        & 0.788 & 0.841 & 0.879 & 0.917 & 0.962 & 0.962 & 0.962 & 0.902 & 369 \\
    0.5        & 0.795 & 0.841 & 0.909 & 0.947 & 0.947 & 0.970 & 0.970 & 0.911 & 409 \\
    \hline
  \end{tabular}
  \label{tab:ratio}
\end{table}

In Sub-section \ref{sec_implement}, we applied a strategy that each batch of training samples was randomly sampled according to a pre-defined positive sampling ratio. Here we evaluated the effects of positive sampling ratios on the lung nodule detection performance of the transformer model.
The results in Table \ref{tab:ratio} show that when the positive sampling ratio was set to 0.2, the result is the best.
Actually, the larger positive sampling ratio the more positive nodules the model tends to predict, as demonstrated in Table \ref{tab:ratio}.
Nevertheless, it can be seen that the model is quit robust to this hyper-parameter as there is no big difference in performance.

\section{Discussions and Conclusion}
In this study, we have adapted the ViT model and unsupervised contrastive learning for lung nodule detection from a CT image volume.
Using neither multi-scale inputs nor assembling techniques, our presented transformer model pretrained in an unsupervised manner has outperformed the state-of-the-art 3D CNN models.
Importantly, we have found that unsupervised representation learning or pretraining on a large-scale dataset can significantly benefit the transformer model, which is scalable, and highly desirable especially when labeled data are scarce.

Our pilot results suggest that for the medical analysis tasks where labeled data are expensive and limited, it is very promising to build a large-scale model, pre-trains it on a related big dataset via  domain-knowledge driven self-supervised, and transfers the learned large-scale prior to benefit down-stream tasks.
Although this study was only focused on CT image representation learning, combining specific imaging modality data with other modalities, such as diagnostic text reports, clinical data, other imaging approaches, etc, has potential to unleash strong power of AI for diagnosis and treatment.

In conclusion, we have presented an adapted 3D ViT model pretrained via region-based contrastive learning for lung nodule detection. Specifically, we have introduced how to adapt the generic transformer model for lung nodule detection.
To make the transformer model work well on a relatively small labeled dataset, we have introduced a self-learning method leveraging public CT data.
The comparative results have demonstrated the superiority of the presented approach over the state of the art 3D CNN baselines.
These findings suggest a promising direction to improve the CAD systems via deep learning.

\section{Acknowledgement}
This work was supported in part by NIH/NCI under Award numbers R01CA233888, R01CA237267, R21CA264772, and NIH/NIBIB under Award numbers R01EB026646, R01HL151561, R01EB031102.

\bibliographystyle{dcu}
\bibliography{reference}

\end{document}